\documentclass{article}

\usepackage{microtype}
\usepackage{graphicx}
\usepackage{subfigure}
\usepackage{booktabs} 
\usepackage{multirow}

\usepackage{hyperref}

\usepackage[accepted]{icml2021}

\icmltitlerunning{Evaluation of Saliency-based Explainability Methods}

\begin{document}

\twocolumn[
\icmltitle{Evaluation of Saliency-based Explainability Methods}




\begin{icmlauthorlist}
\icmlauthor{Sam Zabdiel Sunder Samuel}{rpr}
\icmlauthor{Vidhya Kamakshi}{rpr}
\icmlauthor{Namrata Lodhi}{rpr}
\icmlauthor{Narayanan C Krishnan}{rpr}
\end{icmlauthorlist}

\icmlaffiliation{rpr}{Department of Computer Science and Engineering, Indian Institute of Technology Ropar, Rupnagar, Punjab, India}

\icmlcorrespondingauthor{Sam Zabdiel Sunder Samuel}{2019csm1012@iitrpr.ac.in}

\icmlkeywords{Machine Learning, ICML, XAI, Saliency}

\vskip 0.3in
]



\printAffiliationsAndNotice{}  

\begin{abstract}
A particular class of Explainable AI (XAI) methods
provide saliency maps to highlight part of the image a Convolutional Neural Network (CNN)
model looks at to classify the image as a way to explain its
working. These methods provide an intuitive way for users to
understand predictions made by CNNs. Other than quantitative
computational tests, the vast majority of evidence to highlight
that the methods are valuable is anecdotal. Given that humans
would be the end-users of such methods, we devise three human
subject experiments through which we gauge the effectiveness of
these saliency-based explainability methods.
\end{abstract}

\section{Introduction}
Saliency-based explainability methods provide the image region that CNN models pay attention to make predictions. But there are no specific qualitative metrics in the literature to evaluate the different approaches. This lack of evaluation measure warrants the devising of a system to rate explainability methods.

There have been works to evaluate the different saliency methods. Sanity checks were performed to see if methods were affected by randomization of the model parameters or data labels \cite{adebayo2018sanity}. Sixt et al. \cite{sixt2020explanations} advance a theoretical justification for why many modified backpropagation attributions fail to produce class sensitive saliency maps. Tjoa and Guan \cite{tjoa2020quantifying} perform quantitative analysis of different saliency-based XAI methods, which are shown to be helpful depending on the context. While all quantitative analyses need consideration, a tantamount quality of these XAI methods is to satisfy the cognitive rationales of its users \cite{li2020quantitative_experimental}. A user study has also shown that XAI methods are effective in helping users understand and predict a models output better \cite{alqaraawi2020evaluating}.

There are several desirable qualities of a method that can serve as potential evaluation measures. \textit{We can consider the degree of interpretablity of the explanation. The reliability of explanations is also an important factor to measure the efficacy}. The users of the method should also be able to trust the generated explanations. Considering these desired qualities, we propose three human subject experiments to evaluate the saliency-based explainability methods. We want to evaluate XAI methods against traits such as predictability, reliability, and consistency. These characteristics may not be feasibly measured by quantitative tests, as they are qualitative in nature. Therefore, we conduct human subject experiments to evaluate these methods.

In our first experiment, which \textit{effectively evaluated the predictability of the explanations generated by the methods along with its class-discriminativeness}, GradCAM performed best. In the other two experiments, which \textit{measured how well the explanations from the methods highlight the misclassifications and the consistency in the explanations}, FullGrad \cite{srinivas2019full}, and GradientSHAP \cite{lundberg2017unified} performed well. FullGrad was also deemed consistent in our experiments.

\section{Materials and Methods}
\begin{figure*}[bt]
\centering
\hfill
\subfigure[A question from an \textbf{Experiment \#1} survey. Users are shown the original image and asked to choose from top-4 saliency maps (shown). In the screenshot, the last saliency map is selected which is highlighted by green outer box. \label{subfig:exp1_screen}]{%
\includegraphics[width=\linewidth]{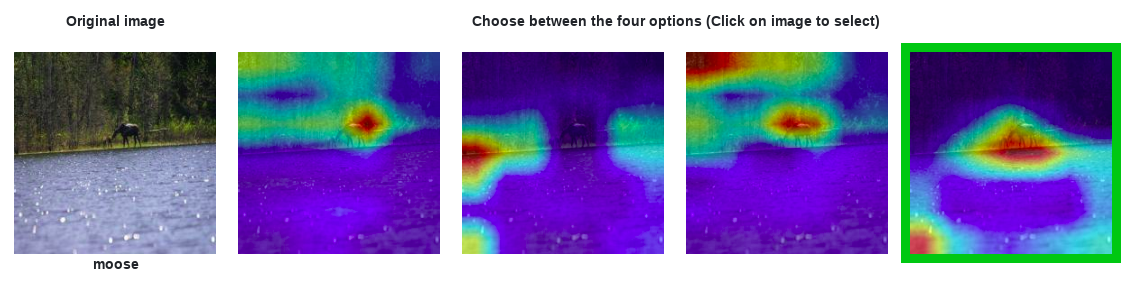}}
\hfill
\subfigure[A question from an \textbf{Experiment \#2} survey. Original image (center) shown along with true class saliency map (center) and predicted class (right). User has to rate on the scale provided below the images the degree to which the saliency maps justify the misclassifications. \label{subfig:exp2_screen}]{%
\includegraphics[width=0.46\linewidth]{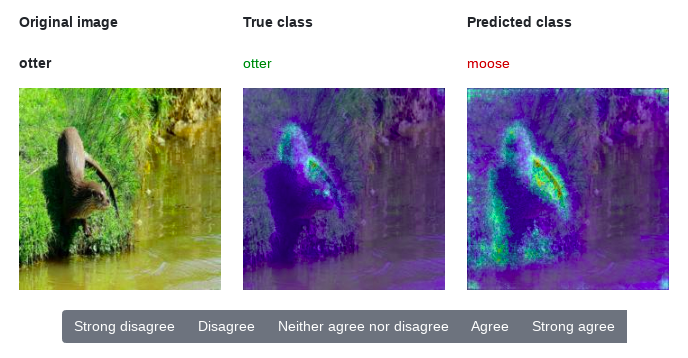}}
\hfill
\subfigure[A question from an \textbf{Experiment \#3} survey. Five original and corresponding saliency maps shown. User has to rate on the scale provided below the images teh degree of consistency in the saliency maps of different images of same bird.\label{subfig:exp3_screen}]{%
\includegraphics[width=0.46\linewidth]{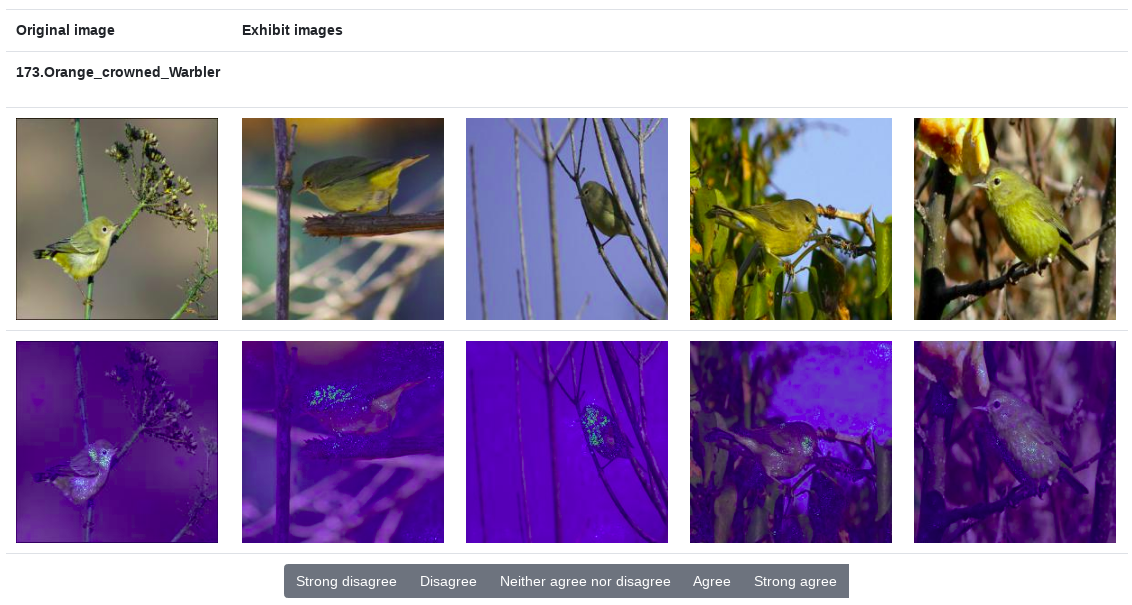}}
\caption{[Best viewed in color] Screenshots of saliency heatmap legend and each experiment's example question from the surveys.}
\label{fig:screengrabs}
\end{figure*}

We propose three human subject experiments to evaluate the efficacy of saliency-based explainability methods. The first two experiments use a subset of Animals with Attributes 2 (AWA2) dataset \cite{xian2018zero} having 20 animal classes. The third experiment uses Caltech-UCSD Birds (CUB) dataset \cite{WahCUB_200_2011} having 200 species of birds. We use VGG-16 \cite{simonyan2014very} as baseline architecture to train our models for the AWA2 and CUB-200 datasets to an accuracy of 96.6\% and 73.0\%, respectively. The experiments are conducted in the form of surveys. The screenshots for the experiments are provided in Figure~\ref{fig:screengrabs}. Each survey had 24 questions. For each survey we received around 20-25 responses. The participants of the experiment were familiar with Artificial Intelligence and Machine Learning but had no prior experience in XAI.

\subsection{Saliency-based XAI methods}
The saliency-based XAI methods (along with their abbreviated form) considered for the experiments are the following:
\begin{enumerate}
    \item Saliency \cite{simonyan2013deep}: Generates saliency map on the basis of gradients at the input layer resulting in a heat map. The heat map is at a fine-grained pixel-level as we have gradients for each pixel.
    \item GradCAM \cite{gradCAM}: Converts the gradients at the final convolutional layer of the CNN into a heat map highlighting important regions. The method saliency maps are at a broad region-level.
    \item Contrastive Excitation Backprop (c-EBP) \cite{zhang2018top}: It results in region-level saliency maps. Attributions are backpropagated in place of gradients. Attribution for the desired class is taken as positive, whereas for all the other classes is taken as negative, resulting in a contrastive explanation. 
    \item FullGrad \cite{srinivas2019full}: The approach also utilizes the gradients for the bias at each intermediate convolutional layer resulting in region-level heat maps that also have granular focus.
    \item Integrated Gradients (IG) \cite{sundararajan2017axiomatic}: The approach uses an axiomatic formulation to satisfy completeness, a desirable quality for the generated explanations. Gradients at the input layer result in pixel-level saliency maps.
    \item DeepLIFT \cite{shrikumar2017learning}: A contribution score based on the difference between the output of the original input and a reference point is backpropagated to each neuron. The approach also considers the positive and negative influences of neurons separately. DeepLIFT generates pixel-level saliency maps.
    \item Gradient SHAP (GradSHAP) \cite{lundberg2017unified}: Inspired from game-theoretic principles, this approach uses Shapley values to gauge importance of each pixel.
    \item SmoothGrad (SG) \cite{smilkov2017smoothgrad}: This variant also provides pixel-level saliency maps by random sampling the inputs by adding the Gaussian noise to the original input. Multiple saliency maps are generated each random-sampled input, and the average of these maps results in the final SmoothGrad saliency map. The vanilla SmoothGrad considers Saliency \cite{simonyan2013deep} as its base XAI method.
    \item Integrated Gradients - SmoothGrad (IG-SG): SmoothGrad applied using Integrated Gradients as the base XAI method to provide pixel-level saliency maps.
\end{enumerate}

\subsection{Experiment \#1-Predictability}\label{exp1_description}
The first experiment's goal is to measure the predictability of the explanation methods. This measure also captures the class-discriminativeness of the generated explanations.
Participants are presented with the explanations for the top-4 predicted class for an image, including the explanation for the ground truth class. Of the four saliency maps, the participants are asked to select the saliency map corresponding to the ground truth class. Ideally, the saliency map of the ground truth class would either demarcate the object of interest in the image from the background or highlight obvious discriminating regions. The average accuracy of the subjects correctly selecting the true class's saliency map would be the explanation method's score. A higher score in this experiment means that the participants can predict the method's outcome among the few options. The measure also incorporates class-discriminativeness as the subjects would only easily select the true class's saliency map if the maps are discriminative.

\subsection{Experiment \#2-Reliability}\label{exp2_description}
The second experiment measures the reliability of the methods. Justifiable explanations even for misclassifications make the method reliable. For this experiment, we show the participants the explanations for misclassified images, i.e., misclassified (predicted) class vs true class (ground truth) saliency maps. The subjects rate on a Likert scale the degree to which they agree that the explanations are justified for the misclassified image. The Likert scale (and corresponding values in parenthesis) that users could select are: Strongly disagree (1), Disagree (2), Neither agree nor disagree (3), Agree (4) and Strongly agree (5). The average rating by the subjects reflects the reliability of each method.

\subsection{Experiment \#3-Consistency}\label{exp3_description}
The third experiment aims to evaluate the consistency of the saliency-based explainability methods. By consistency, we mean that the explanation method for similar images would highlight similar regions (having same patterns) of the image. In the third experiment, we show the user several saliency maps of correctly predicted images belonging to the same class. We ask the participant to rate on a Likert scale (similar to one in experiment 2) the degree to which they agree that the saliency maps of different images of the bird (from the CUB dataset) focuses on the same or similar part(s) of the bird. This experiment is particularly beneficial in the case of the fine-grained image classification task.


\section{Results}
\begin{figure}[!t]
    \centerline{\includegraphics[width=0.96\linewidth]{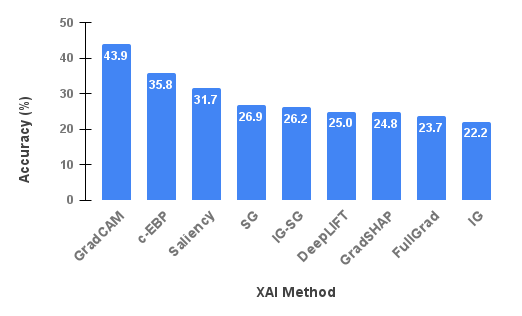}}
  \caption{Experiment \#1-Predictability accuracy}
  \label{fig:exp1_accuracy} 
\end{figure}

We show the accuracies for the task outlined in experiment \#1 in Figure~\ref{fig:exp1_accuracy}. GradCAM has the highest accuracy of 43.9\% among all the XAI methods. Excitation Backprop and Saliency follow this with accuracies of 35.8\% and 31.7\%, respectively. There is a substantial difference in the score of the top-2 performing methods. Integrated-Gradients and FullGrad are the worst-performing XAI methods for the task, with accuracies of 22.2\% and 23.7\%. 

Table~\ref{exp2n3_table} shows the ranks of the XAI methods based on the mean values on the Likert scale  \cite{boone2012analyzing} \cite{sullivan2013analyzing} for experiment \#2. GradSHAP and FullGrad have the highest mean, 3.38 and 3.36, respectively. Saliency and GradCAM again feature in the top four with means of 3.26 and 3.22, respectively.




FullGrad has a Likert value mean of 4.51 in experiment \#3 as reflected in Table~\ref{exp2n3_table}. This implies that there is a near-unanimous consensus that the explanations provided by FullGrad are consistent for the VGG-16 model trained on the CUB-200 dataset. Other methods are also closely behind except GradCAM and DeepLIFT, which have mean values of 3.22 and 3.36, respectively. The frequency for ``Strongly agree" in the case of FullGrad defies trends to stand out as the highest among the Likert items of the method. It covers 61.28\% of the total responses for the method. The second highest was for c-EBP having 29.6\%.

\begin{table}[!tb]
\centering
\caption{Likert values mean for Experiment \#2 and \#3}
\vskip 0.15in
\begin{center}
\begin{small}
\begin{tabular}{|c|c|c|}
\hline
\multirow{3}{4em}{\textbf{XAI Method}} & \multicolumn{2}{|c|}{\textbf{Mean$\pm$SD\textsuperscript{Rank}}} \\
\cline{2-3}
 &  \multirow{2}{8em}{\textbf{Experiment \#2 Reliability}} & \multirow{2}{8em}{\textbf{Experiment \#3 Consistency}} \\
 &&\\
\hline
FullGrad & 3.36$\pm1.13^\textbf{2}$ & 4.51$\pm0.71^\textbf{1}$\\
c-EBP & 3.15$\pm1.27^7$ & 3.92$\pm1.00^\textbf{2}$\\
GradSHAP & 3.38$\pm1.08^\textbf{1}$ & 3.90$\pm0.87^\textbf{3}$\\
IG & 3.11$\pm1.16^9$ & 3.82$\pm0.84^4$\\
SG & 3.18$\pm1.14^6$ & 3.80$\pm1.06^5$\\
Saliency & 3.26$\pm1.16^\textbf{3}$ & 3.79$\pm0.96^6$\\
IG-SG & 3.19$\pm1.12^5$ & 3.71$\pm0.95^7$\\
DeepLIFT & 3.13$\pm1.67^8$ & 3.36$\pm1.17^8$\\
GradCAM & 3.22$\pm1.18^4$ & 3.22$\pm1.13^9$\\
\hline
\end{tabular}
\label{exp2n3_table}
\end{small}
\end{center}
\vskip -0.1in
\end{table}

\section{Discussion}

We considered the AWA2 dataset for the first two tasks as it did not have too much variability considering that the backgrounds and even animals in many cases were similar, like bobcat and tiger, but the classes were also not too similar. The use of the CUB-200 dataset is also significant for the third experiment. The images capture the birds in very few ways, like either perched or flying. This helps us to look at the consistency of the saliency maps better, as, in the case of the AWA2 dataset, the animals are captured from various angles, distances, environments, and various positions. 

Experiment \#1 shows that although saliency maps are predictable, GradCAM scoring 43.9\% accuracy despite no context being provided, the utility of its predictability cannot be ascertained. Among the methods, GradCAM and Excitation Backprop having the highest accuracy can also be attributed to the fact that GradCAM has highlights large regions of the image where boundaries may not be well-demarcated \cite{srinivas2019full}. Excitation Backprop is also helped by the same fact and also that we are using the contrastive variant, which gives well discriminative saliency maps. All the other methods provide pixel-level saliency maps except FullGrad, although it also provides a finely detailed saliency map. The fineness of the details of the saliency maps may have had counterproductive effects. Added by the fact that the model itself tries to discriminate between classes by specific features like object parts, textures, colours, even backgrounds \cite{gonzalez2018semantic} \cite{bau2017network} and not strictly focuses entirely on the object of interest.

The effectiveness of FullGrad is also not surprising in the second and third experiment due to the same reasons. It has a mix of finely detailed and broad region covering saliency maps like GradCAM and other pixel-level methods. FullGrad also has a theoretical foundation that credits both local and global attribution.GadientSHAP uses Shapley values, to which its success can be attributed. 

\begin{figure}[tb]
   \centerline{\includegraphics[width=0.9\linewidth]{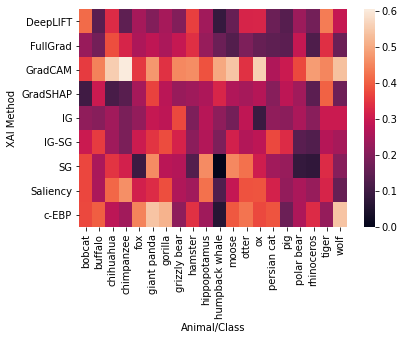}}
  \caption{[Best viewed in color] Experiment \#1 XAI Method vs. Class Heatmap of mean accuracies (as fractions).}
  \label{exp1_heatmap} 
  \vskip -0.2in
\end{figure}

\begin{figure}[tb]
    \centerline{\includegraphics[width=0.9\linewidth]{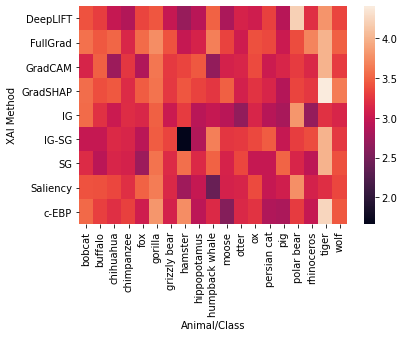}}
  \caption{[Best viewed in color] Experiment \#2 XAI Method vs. Class Heatmap of mean of Likert Value.}
  \label{exp2_heatmap} 
  \vskip -0.2in
\end{figure}

Fig.~\ref{exp1_heatmap} shows the heatmap of the mean accuracies in Experiment 1 for the methods against each class. We want to bring attention to the fact that some classes exhibit better accuracy overall, indicated by a lighter tone in its column relative to other classes, like giant panda, gorilla and tiger. The opposite can be seen in the case of humpback whale, polar bear rhinoceros, which have relatively darker columns. This similar phenomenon is observed in Fig.~\ref{exp2_heatmap}, which shows the heatmap of the mean Likert values in Experiment 2 for the methods against each class. The classes gorilla and tiger are relatively lighter contrasted with the classes humpback whale, pig and rhinoceros, which are relatively darker, as seen by their corresponding columns. This phenomenon suggests that explainability differ among individual classes.

\section{Conclusion}
We conducted human subject experiments to gauge the effectiveness of saliency-based XAI methods, which provide saliency maps as explanations for predictions made by CNN models. The experiments were designed to capture whether the XAI methods had desirable qualities like predictability, reliability, and consistency. Our work has found the majority of these methods to possess the desirable qualities, but not to a high degree of proficiency. Some methods have come out on top, like GradientSHAP and FullGrad. FullGrad was also rated highly consistent nearly unanimously by the human subjects. The benefits of XAI methods to humans can only be quantified by how well it satisfies its human users. As suggested by the experiments, a human-centric approach to XAI might assist to bridge the vast gap that exists between the explanations and their utility.
\bibliography{example_paper}
\bibliographystyle{icml2021}

\end{document}